\documentclass[submission,copyright,creativecommons]{eptcs}
\providecommand{\event}{ICLP DC 2021}

\usepackage{xspace}
\usepackage{graphicx}

\def\mapf{{MAPF}\xspace}
\def\mmapf{{mMAPF}\xspace}
\def\dmapf{{\bf D-MAPF}\xspace}

\title{Flexible and Explainable Solutions for \\ Multi-Agent Path Finding Problems}
\author{Aysu Bogatarkan
\institute{Faculty of Engineering and Natural Sciences, Sabanci University, Istanbul, Turkey}
\email{aysubogatarkan@sabanciuniv.edu}}

\def\titlerunning{Flexible and Explainable Solutions for MAPF}
\def\authorrunning{A. Bogatarkan}
\begin{document}
\maketitle
\begin{abstract}
The multi-agent path finding (\mapf) problem is a combinatorial search problem that aims at finding paths for multiple agents (e.g., robots) in an environment (e.g., an autonomous warehouse) such that no two agents collide with each other, and subject to some constraints on the lengths of paths. The real-world applications of \mapf require flexibility (e.g., solving variations of \mapf) as well as explainability. In this study, both of these challenges are addressed and some flexible and explainable solutions for \mapf and its variants are introduced. 
\end{abstract}

\section{Introduction}

Artificial Intelligence (AI) applications are being used widely among people with different background and interests. For these applications to be successful, two of the important features (and challenges) needed by AI methods are flexibility and explainability. A flexible AI method developed to solve a problem can accommodate variations of the problem, and thus can be used to investigate different options for a better understanding. An explainable AI method can provide answers to queries about the (in)feasibility and the optimality of solutions. One of the well-studied problems in AI that necessitates solutions for these two challenges is the multi-agent path finding (\mapf) problem.

\mapf problem aims to find plans for multiple agents in an environment without colliding with each other or obstacles, subject to some constraints on the maximum or the total plan length. Optimal solutions for \mapf can be found by optimizing the makespan of the whole plan or the total plan length of agents. These optimization functions can be extended according to the needs of an application. While single-agent shortest pathfinding can be solved in polynomial time~\cite{Dijkstra59}, \mapf with constraints on plan lengths is intractable~\cite{RatnerW86}. \mapf has been studied in various domains, such as robotics~\cite{LeeY09}, traffic control~\cite{DresnerS08}, video games~\cite{StandleyK11}, autonomous aircraft towing vehicles~\cite{MorrisPLMMKK16},  and autonomous warehouse systems~\cite{WurmanDM08}.

This study focuses on finding flexible frameworks for \mapf and its variants and an explainable framework for one of these variants of \mapf. In all of these solutions, Answer Set Programming (ASP)~\cite{MarekT99,Niemelae99,Lifschitz02,BrewkaET11,BrewkaEL16}---a logic programming paradigm based on answer sets~\cite{gelfondL91,gelfond1988stable} is utilized. 

For some real-world applications, being able to solve \mapf problem may not be enough to address all challenges or the realistic conditions of the application. For instance, in real-world automated warehouses, the robots’ battery levels change as they travel around and it may be necessary for them to be charged to complete their tasks. Furthermore, some parts of the warehouses (for instance with human occupants or tight passages) may require robots to move slowly to ensure safety. In a warehouse that is not completely autonomous, some changes may occur during the execution of a plan: existing agents may leave the environment, or new agents may be included in the team with new tasks, existing obstacles may be removed from the environment or moved to some other location in the environment. One aim of our study is to address these issues with with flexible frameworks in the spirit of elaboration tolerance~\cite{mccarthy98}.

We also investigate the challenge of explainability for one of these problems, in particular, considering queries about the (in)feasibility and the optimality of solutions, as well as queries about the observations about these solutions. Given a solution for a general variation of \mapf, our explainable framework is able to explain infeasibility or nonoptimality of this solution, confirm its feasibility and suggest alternatives for the solution, and provide explanations for some queries.

\section{Flexible Solutions for \mapf: Current Status}

Let us describe our contributions and current status of our research on flexible frameworks for \mapf.

\subsection{Multi-Modal Multi-Agent Path Finding}

In real-world automated warehouses, the robots’ battery levels change as they move around, and, in some parts of these warehouses, due to presence of humans or tight passages, the robots may need to move slowly to ensure safety. Hence, to be able to handle more realistic autonomous warehouse scenarios, a mathematical model general enough to handle multi-modal transportation conditions and multi-objective optimizations is needed. Furthermore, the computational framework is required to be flexible such that a large set of variations of \mapf problems can be addressed. Motivated by these challenges, we mathematically modelled a general version of \mapf (called \mmapf -- multi-modal \mapf with resources) as a rich graph problem and introduced a flexible method to solve \mmapf declaratively, using ASP~\cite{bogatarkan2020multi}. We have implemented our framework using the ASP solver Clingo~\cite{gebser2011clingo}. Our method can handle the following variations of \mapf: \textit{multi-objective optimization}, \textit{waypoints}, \textit{resource constraints} and \textit{multi-modal transportation}.

\begin{figure}[ht]
    \centering
    \includegraphics[scale=0.38]{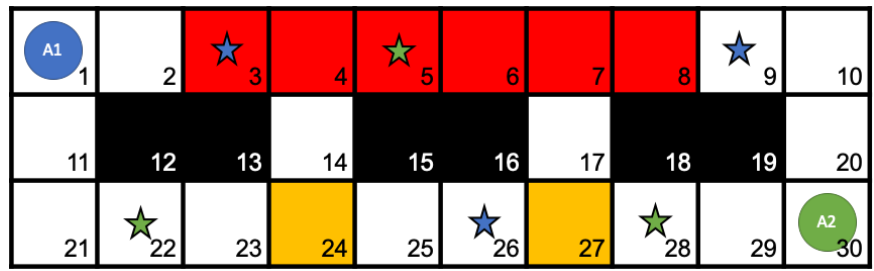}
    \caption{An \mmapf instance. Initially, robot A1 is located at cell 1 and robot A2 is located at cell 30. The goal of A1 is to deliver some items to cell 30, while the goal of A2 is to deliver some items to cell 1. Each robot’s waypoints are shown by stars with the same color as that robot.}
    \label{fig:mmapf}
\end{figure}

Figure~\ref{fig:mmapf} shows an example \mmapf scenario introduced in our earlier studies~\cite{bogatarkan2020multi}. In this example, the warehouse consists of three shelf units denoted as obstacles (black cells). The charging stations are located at cells 24 and 27 highlighted by yellow, and the corridor where the agents should go slowly, covers cells 3--8 highlighted by red. In this area, it takes 2 timesteps to move from one cell to its neighbor. The maximum battery level is set to 10. Robot A1 starts with an initial battery level of 10, while robot A2 has an initial battery level of 8. Each robot wants to collect some items on its way (denoted by stars that match the color of the relevant robot) and to deliver these items to the opposite corner of the warehouse. Therefore, eventually, they want to swap their places.

\begin{figure}[ht]
    \centering
    \includegraphics[]{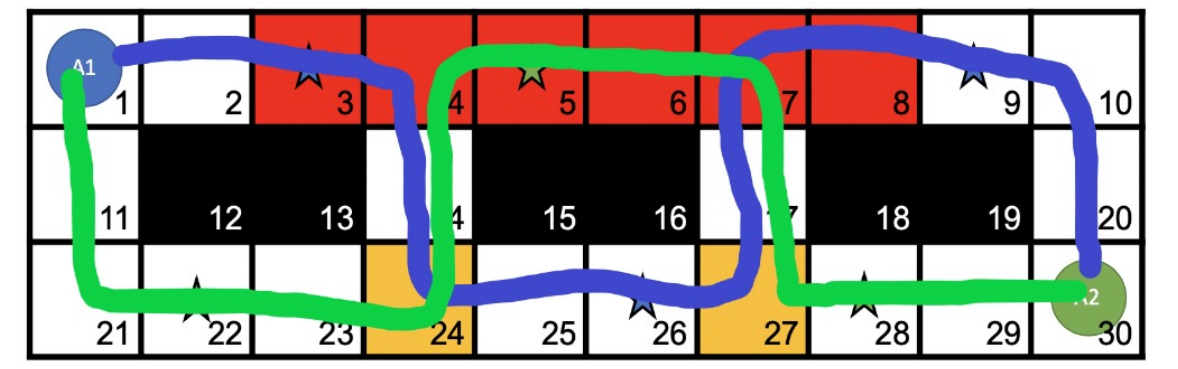}
    \caption{A solution for the \mmapf instance shown in Figure~\ref{fig:mmapf}. A1 follows the blue path while A2 follows the green path to reach their goals, ensuring they collect items at their waypoints.}
    \label{fig:mmapf_sol}
\end{figure}

A solution for this \mmapf instance is shown by the colored paths in Figure~\ref{fig:mmapf_sol}. Note that each robot visits its waypoints on the way to the goal. The traversals of these paths (i.e. location of each robot in each timestep) are shown in Table~\ref{tab:mmapf_sol}. The traversal of each slow edge (e.g., A2 moving from cell 6 to 5) takes two timesteps. The battery levels of the robots are also shown in this table. The battery level decreases at each time step, unless a robot is at a charging station. The battery level gets to its maximum when robots decide to charge while at a charging station. For instance, A2's battery level increases to 10 when the charging station at cell 27 is visited.

\begin{table}[ht]
\label{tab:mmapf_sol}
\centering
\resizebox{\columnwidth}{!}{
    \begin{tabular}{|l||c|c|c|c|c|c|c|c|c|c|c|c|c|c|c|c|c|c|c|}
        \hline
        Time & 0 & 1 & 2 & 3 & 4 & 5 & 6 & 7 & 8 & 9 & 10 & 11 & 12 & 13 & 14 & 15 & 16 & 17 & 18 \\ \hline
        $A1$ Location & 1 & 2 & 3 & transit & 4 & 14 & 24 & 25 & 26 & 27 & 17 & 7 & transit & 8 & 9 & 10 & 20 & 30 & -- \\ \hline
        $A1$ Battery & 10 & 9 & 8 & 7 & 6 & 5 & 4 & 3 & 2 & 1 & 10 & 9 & 8 & 7 & 6 & 5 & 4 & 3 & -- \\ \hline
        $A2$ Location & 30 & 29 & 28 & 27 & 17 & 7 & transit & 6 & transit & 5 & transit & 4 & 14 & 24 & 23 & 22 & 21 & 11 & 1 \\ \hline
        $A2$ Battery & 8 & 7 & 6 & 5 & 10 & 9 & 8 & 7 & 6 & 5 & 4 & 3 & 2 & 1 & 10 & 9 & 8 & 7 & 6 \\ \hline
    \end{tabular}
    }

\end{table}

Details of the mathematical model and the ASP formulation can be found in our paper, \textit{Multi-Modal Multi-Agent Path Finding with Optimal Resource Utilization}~\cite{bogatarkan2020multi}.

\subsection{Dynamic Multi-Agent Path Finding}

Another flexible framework introduced was an algorithm for \mapf in dynamic environments. When changes occur in a dynamic environment, such as obstacles being removed or moved, existing agents leaving the environment or new agents being added to the team, the aim is to find a new solution for the new team of agents in the modified environment. We call this problem Dynamic Multi-Agent Path Finding Problem (\dmapf)~\cite{bogatarkanPE19}.

One of the possible solutions for \dmapf is replanning: consider a new \mapf instance defined by the current locations and goal locations of both the existing and the new agents, and the updated environment, and compute a solution for this instance. Although replanning finds a solution, if one exists, it does not re-use the plans of the existing agents and may not be computationally efficient.

With this motivation, we proposed a novel method to solve \dmapf, using Answer Set Programming (ASP) . The main idea (and novelty) of this method is, instead of replanning for all the agents right away, to {\em revise and augment} the existing \mapf solution: ({\em revise}) try to schedule the waiting times of existing agents as they traverse the rest of their paths, ({\em augment}) while computing paths for the new agents within a given makespan (i.e., the length of the plan). In this way, the paths for the existing agents can be re-used as part of the new plan. We have implemented our framework using Python and the ASP solver Clingo. In our experimental evaluations, we observed that the re-use of plans as proposed by our method improves the computational efficiency in timings significantly compared to replanning. 

\begin{figure}[ht]
	\centering
	\rotatebox{0}{\resizebox{0.20\columnwidth}{!}{\includegraphics{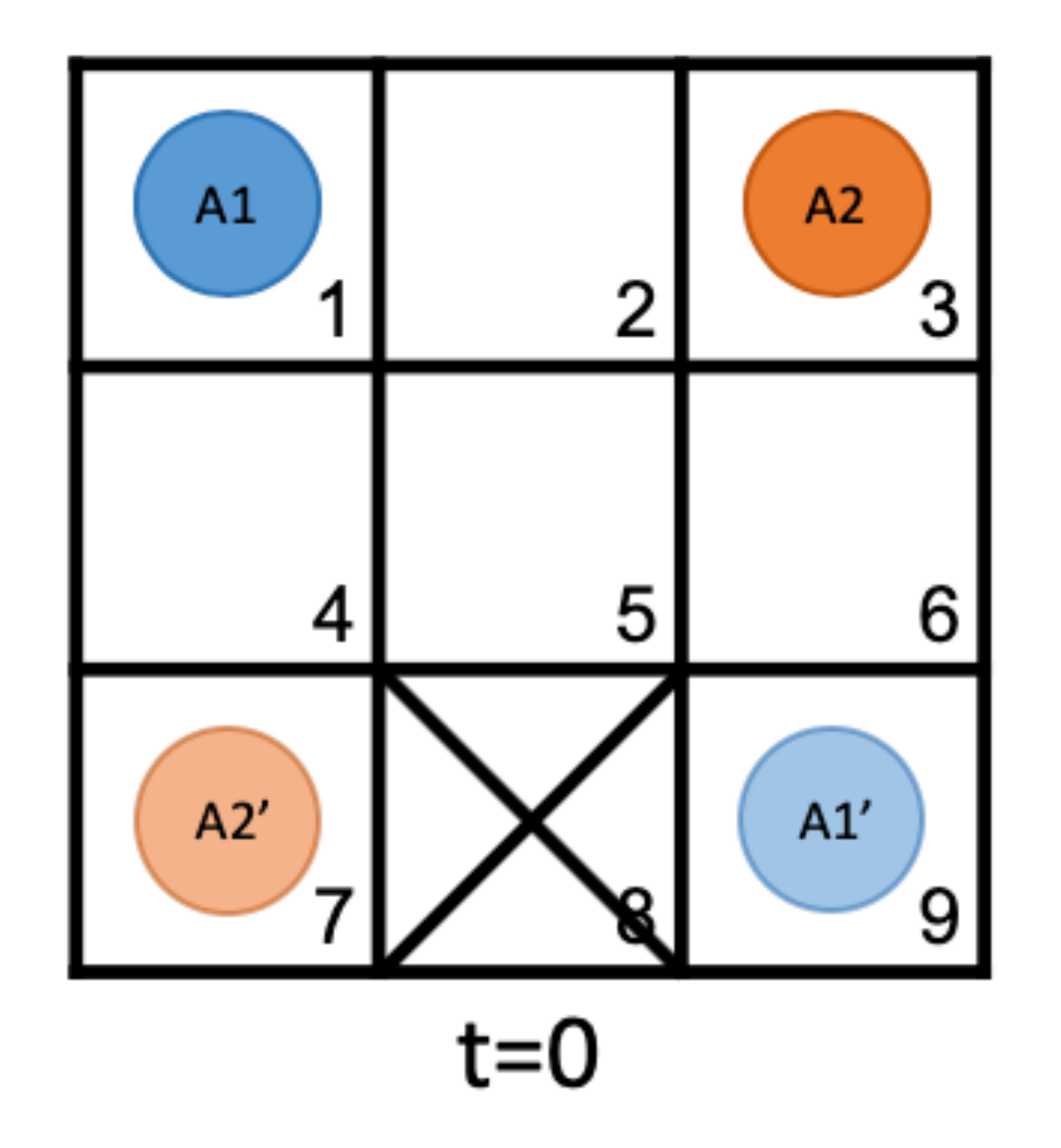}}}
	\vspace{-1\baselineskip}
	\caption{Initially, two agents are at $A1$ and $A2$; their goals are to reach $A1'$ and $A2'$.}
   \vspace{0.5\baselineskip}
	\label{fig:dmapf0}
\end{figure}

\begin{figure}[ht]
	\centering
	\rotatebox{0}{\resizebox{0.85\columnwidth}{!}{\includegraphics{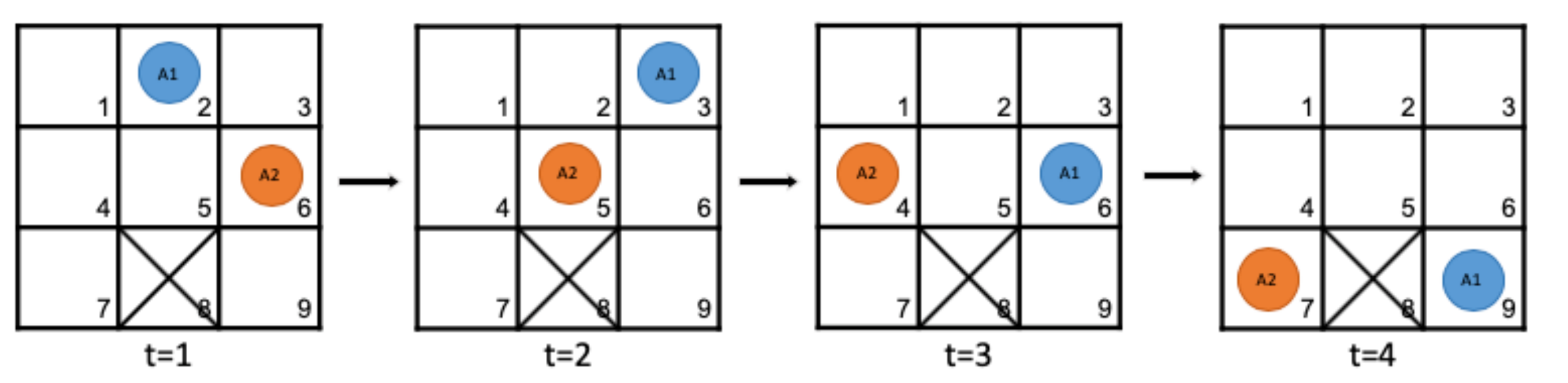}}}
	\vspace{-\baselineskip}
	\caption{A solution to \mapf instance described in Fig.~\ref{fig:dmapf0}: $A1$ starts at 1, moves to 2, 3, 6, and reaches 9; and $A2$ starts at 3, moves to 6, 5, 4, reaches 7.}
    \vspace{.5\baselineskip}
	\label{fig:dmapf1}
\end{figure}

\begin{figure}[!ht]
	\centering
	\rotatebox{0}{\resizebox{0.85\columnwidth}{!}{\includegraphics{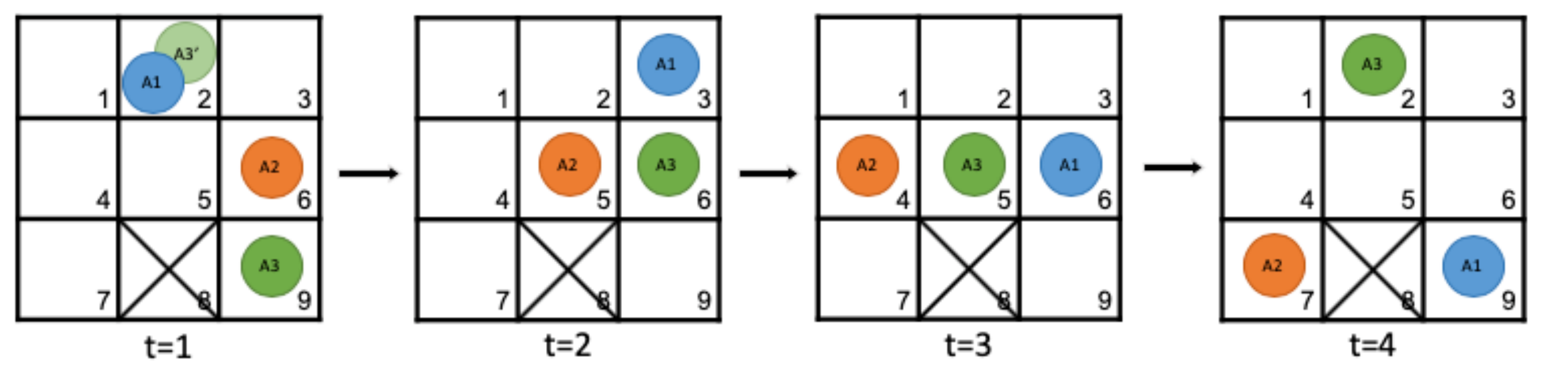}}}
	\vspace{-\baselineskip}
	\caption{While executing the plan shown in Fig.~\ref{fig:dmapf1}, at time step $t=1$, another agent $A3$ joins the team with goal $A3'$. A solution to this \dmapf instance is computed, and a path for $A3$ is found, as shown above: $A1$ is at 2, moves to 3, 6, 9; $A2$ is at 6, moves to 5, 4, 7; $A3$ starts at 9, moves to 6, 5, 2.}
   \vspace{.5\baselineskip}
	\label{fig:dmapf2}
\end{figure}

\begin{figure}[!ht]
	\centering
	\rotatebox{0}{\resizebox{0.85\columnwidth}{!}{\includegraphics{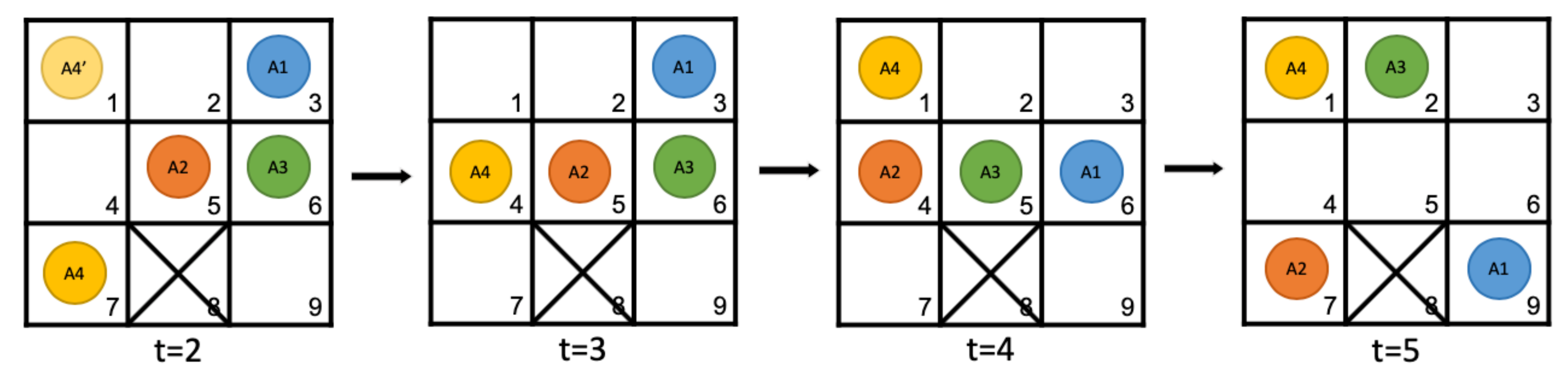}}}
	\vspace{-\baselineskip}
	\caption{While executing the plan shown in Fig.~\ref{fig:dmapf2}, at time step $t=2$, another agent $A4$ joins the team with goal $A4'$ . A solution to this \dmapf instance is computed, no solution found makespan $t=4$, so the makespan is increased by 1. The new solution with makespan $t=5$ is found by scheduling the waiting times of $A1$, $A2$ and $A3$, and by computing a plan for $A4$, as shown above: $A1$ is at 3, waits at 3, then moves to 6,9; $A2$ is at 5, waits at 5, then moves to 4,7; $A3$ is at 6, waits at 6, then moves to 5,2; $A4$ starts at 7, moves to 4, 1, and waits at 1.}
	\label{fig:dmapf3}
\end{figure}

Figures~\ref{fig:dmapf0}--\ref{fig:dmapf3} present an example (from our earlier studies~\cite{bogatarkanPE19}) to illustrate the overall idea of our \dmapf algorithm.


The problem definition, ASP formulation, algorithm and experimental evaluations are described in detail in our paper \textit{A Declarative Method for Dynamic Multi-Agent Path Finding}~\cite{bogatarkanPE19}.

\section{Explainable Solutions for \mapf: Current Status}

We also investigate the challenge of explainability for \mmapf problems, considering queries about the (in)feasibility and the optimality of solutions, along with queries about observations about these solutions. For instance, suppose that an \mmapf solution is being executed in a warehouse and an engineer in this warehouse would like to check whether some modifications of this \mmapf solution would still be feasible or not.

\begin{itemize}
\item {\em Explaining infeasibility or nonoptimality.} Suppose that the modified solution is found infeasible, then, an explanation regarding infeasibility of this modified solution could be ``due to collisions with obstacles or other robots'' or ``due to low battery-level''. An explanation regarding nonoptimality of the modified solution could be ``because some more time is needed to complete tasks'' or ``because some more charging is required''.
\item {\em Confirming feasibility and suggesting alternatives.} Suppose that the modified solution is found feasible. Furthermore, a better solution is computed (e.g., where the tasks are completed earlier).  Then, in addition to confirming that the plan is feasible, alternative solutions could be provided to the engineer.
\end{itemize}

In an alternative scenario, suppose that the engineer would like to better understand the \mmapf solution being executed in the warehouse, and asks various queries about it. For such queries, it is useful to generate explanations using counterfactuals.
 \begin{itemize}
 \item {\em Explaining why an agent is waiting too long at a location.} Suppose that the engineer observes that the agent is waiting for a while at some location, and she wants to know why. An explanation could be that ``if the agent does not wait at that location for a while, it will collide with another robot.'' Alternatively, an explanation could be ``actually, there is no need for the agent to wait there so long, but it needs to follow a different itinerary such as ... to complete tasks on time'' or ``actually, there is no need for the agent to wait there so long, but it needs to follow a different itinerary such as ... and will be late a bit.''
\end{itemize}

Such queries and explanations would help the engineer to better understand the strengths and weaknesses of the solution being executed, as well as the limitations of the infrastructure.

With these motivating real life scenarios, we introduced a method to generate explanations for a variety of queries about \mmapf solutions, using the expressive formalism and efficient solvers of ASP~\cite{bogatarkanE20expl}. Our explainable framework is implemented using Python and the ASP solver Clingo.

\medskip

\noindent\textbf{An Example Scenario for a Query About Waiting}

\begin{figure}[ht]
\begin{tabular}{ccc}
\centering
        \resizebox{0.3\textwidth}{!}{\begin{tabular}{c}
        \includegraphics[scale = 0.35]{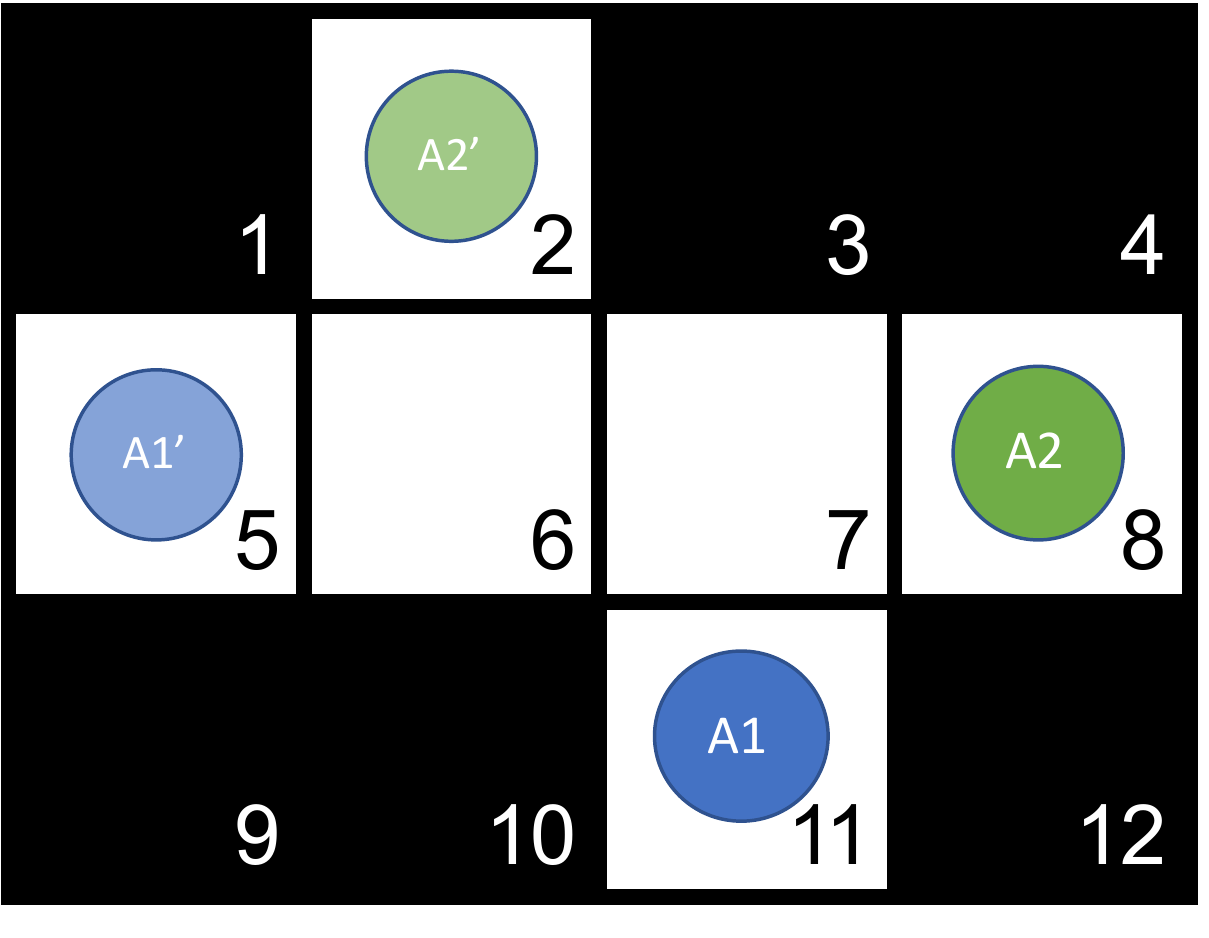}
        \end{tabular}}
        &
        \resizebox{0.3\textwidth}{!}{\begin{tabular}{ccc}
        \hline\hline
        Time & A1 Location & A2 Location \\ \hline
        0 & 11 & 8 \\
        1 & 7 & 8 \\
        2 & 6 & 7 \\
        3 & 5 & 6 \\
        4 & - & 2 \\  \hline\hline
        \end{tabular}}
        &
        \resizebox{0.3\textwidth}{!}{\begin{tabular}{ccc}
        \hline\hline
        Time & A1 Location & A2 Location \\ \hline 
        0 & 11 & 8 \\
        1 & 11 & 7 \\
        2 & 7 & 6 \\
        3 & 6 & 2 \\
        4 & 5 & - \\ \hline\hline
        \end{tabular}}

        \\
        (a) & (b) & (c)  \\
        \end{tabular}
            \vspace{-\baselineskip}
        \caption{\footnotesize(a) A1 and A2 denote the initial positions of Robots~1 and~2; A1$'$ and A2$'$ denote goal locations. Cell~7 is a waypoint for both robots, (b) and (c) are two optimal plans for this instance.}
        \label{fig:expl_fig1}

    \end{figure}

Consider the \mmapf instance (Scenario 1 of~\cite{bogatarkanE20expl}) shown in Figure~\ref{fig:expl_fig1}(a) in a small warehouse, where Robot~1 is initially located at Cell~11 and aims to reach Cell~5, and Robot~2 is initially located at Cell~8 and aims to reach Cell~2. Cell~7 is a waypoint for both robots, and the upper bound on makespan of a plan is 4. For simplicity, suppose that the batteries of the robots are fully charged initially, and sufficient for execution of any given plan, and that all edges have the normal mode of transportation.

About Plan~1 described in Figure~\ref{fig:expl_fig1}(b), suppose that an engineer asks the following query:
\begin{quote}
``Why does Robot~2 wait at Cell~8 (at any time)?''
\end{quote}

Our algorithm checks whether there is an optimal plan where Robot~2 does not have to wait initially.

Once an alternative solution, Plan~2 described in Figure~\ref{fig:expl_fig1}(c), is found, our algorithm presents the following explanation: \begin{quote}
``Actually, Robot~2 does not have to wait at Cell~8 from time step 0 to 2. Here is an alternative optimal plan: Plan~2...''
\end{quote}
\medskip
\noindent\textbf{An Example Scenario for Nonexistence of Solutions Query}
\begin{figure}[ht]
    \centering
    \includegraphics[scale=0.4]{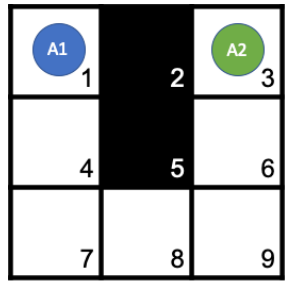} 
    \caption{Robots~1 and~2 are initially located at the given positions on the grid; the goal of each robot is the initial location of the other. Black cells are obstacles.}
    \label{fig:expl_qu}
\end{figure}

Consider the instance (Scenario 6 of~\cite{bogatarkanE20expl})  in Figure~\ref{fig:expl_qu}(b). Robots~1--2 are initially located at $1$ and $3$; their goals are at $3$ and $1$, respectively. Black cells denote obstacles. For simplicity, we assume that their batteries are initially fully charged and enough to traverse all of their paths, all edges have normal mode of transportation, and the only waypoint of each robot is located at its initial position.

There is no solution for this \mmapf instance. Suppose that an engineer wants to find out the reason and asks the following query:
\begin{quote}
``Why does not the instance have a solution?''
\end{quote}

\noindent The algorithm generates the following explanation:
\begin{quote}
``There is no solution because Robot~1 and Robot~2 collide at Cell~8 at time step~3.''
\end{quote}

\noindent The algorithm further generates the following explanation:
\begin{quote}
``There is no solution because Robot~2 collides with the obstacle at Cell~2 at time step~1; this suggests removing this obstacle.''
\end{quote}

Details of the algorithm and the ASP formulations, together with more examples and experimental evaluations are presented in our paper, \textit{Explanation Generation for Multi-Modal Multi-Agent Path Finding with Optimal Resource Utilization using Answer Set Programming}~\cite{bogatarkanE20expl}.

\section{Discussions and Future Work}

We have introduced flexible and explainable frameworks for \mapf problem, addressing different challenges. While \mmapf focuses on a flexible framework considering resource optimization and multi-modality, \dmapf considers dynamic changes in the environment. In addition to these frameworks, an explainable framework for \mmapf is also introduced, generating explanations about feasibility and optimality about given \mmapf solutions along with some observations or suggestions, via queries and counterfactual reasoning. For all these frameworks, we contributed the literature with publications. We have also developed an educational multi-robot system to apply \mapf with robots~\cite{tork2019}.

Currently, we are working on applying these frameworks on some real-world applications in collaboration with a logistics company. We are also working on improving our methods introduced for \dmapf~\cite{bogatarkanPE19} and \mmapf~\cite{bogatarkan2020multi}, extend their experimental evaluations and we aim to publish journal articles for our improved methods. 

\section*{Acknowledgments}

I would like to thank Prof.~Esra Erdem (Sabanci Universitys, Computer Science and Engineering) and Prof.~Volkan Patoglu (Sabanci University, Mechatronics Engineering) for their supervision in this research, Dr.~Orkunt Sabuncu (TED University, Computer Engineering) and Dr.~Alexander Kleiner (Robert Bosch GmbH, Corporate Research) for useful discussions, and the anonymous reviewers for their comments and suggestions. This work is supported by Tubitak Grant~118E931.

\bibliographystyle{eptcs}
\bibliography{references}
\end{document}